\documentclass[runningheads]{llncs}
\usepackage{graphicx}

\usepackage{tikz}
\usepackage{comment}
\usepackage{amsmath,amssymb} 
\usepackage{color}

\usepackage[accsupp]{axessibility}  

\usepackage{booktabs}
\usepackage{xcolor}
\usepackage{pifont}
\newcommand{\cmark}{\ding{51}}
\newcommand{\xmark}{\ding{55}}
\DeclareMathOperator*{\argmax}{arg\,max}

\def\eg{\emph{e.g.}}
\def\ie{\emph{i.e.}}
\def\etal{\emph{et al.}}

\usepackage[pagebackref,breaklinks,colorlinks,linkcolor=blue,citecolor=blue,urlcolor=blue,bookmarks=false]{hyperref}

\begin{document}
\pagestyle{headings}
\mainmatter
\def\ECCVSubNumber{2568}  

\title{Learning Visibility for Robust Dense Human Body Estimation} 

\titlerunning{VisDB: Visibility-aware Dense Body}
%
\author{
Chun-Han Yao$^1$ 
Jimei Yang$^2$ 
Duygu Ceylan$^2$ 
Yi Zhou$^2$ \\
Yang Zhou$^2$ 
Ming-Hsuan Yang$^{134}$
}

%
\authorrunning{Yao et al.}
%
\institute{
$^1$UC Merced 
$^2$Adobe 
$^3$Google 
$^4$Yonsei University
}
\maketitle

\begin{abstract}
Estimating 3D human pose and shape from 2D images is a crucial yet challenging task.
While prior methods with model-based representations can perform reasonably well on whole-body images, they often fail when parts of the body are occluded or outside the frame.
Moreover, these results usually do not faithfully capture the human silhouettes due to their limited representation power of deformable models (e.g., representing only the naked body).
An alternative approach is to estimate dense vertices of a predefined template body in the image space.
Such representations are effective in localizing vertices within an image but cannot handle out-of-frame body parts.
In this work, we learn dense human body estimation that is robust to partial observations.
We explicitly model the visibility of human joints and vertices in the x, y, and z axes separately.
The visibility in x and y axes help distinguishing out-of-frame cases, and the visibility in depth axis corresponds to occlusions (either self-occlusions or occlusions by other objects).
We obtain pseudo ground-truths of visibility labels from dense UV correspondences and train a neural network to predict visibility along with 3D coordinates.
We show that visibility can serve as 1) an additional signal to resolve depth ordering ambiguities of self-occluded vertices and 2) a regularization term when fitting a human body model to the predictions.
Extensive experiments on multiple 3D human datasets demonstrate that visibility modeling significantly improves the accuracy of human body estimation, especially for partial-body cases.
Our project page with code is at: \url{https://github.com/chhankyao/visdb}.
\end{abstract}


\section{Introduction}
\begin{figure}[t]
    \begin{center}
    \includegraphics[width=0.77\linewidth]{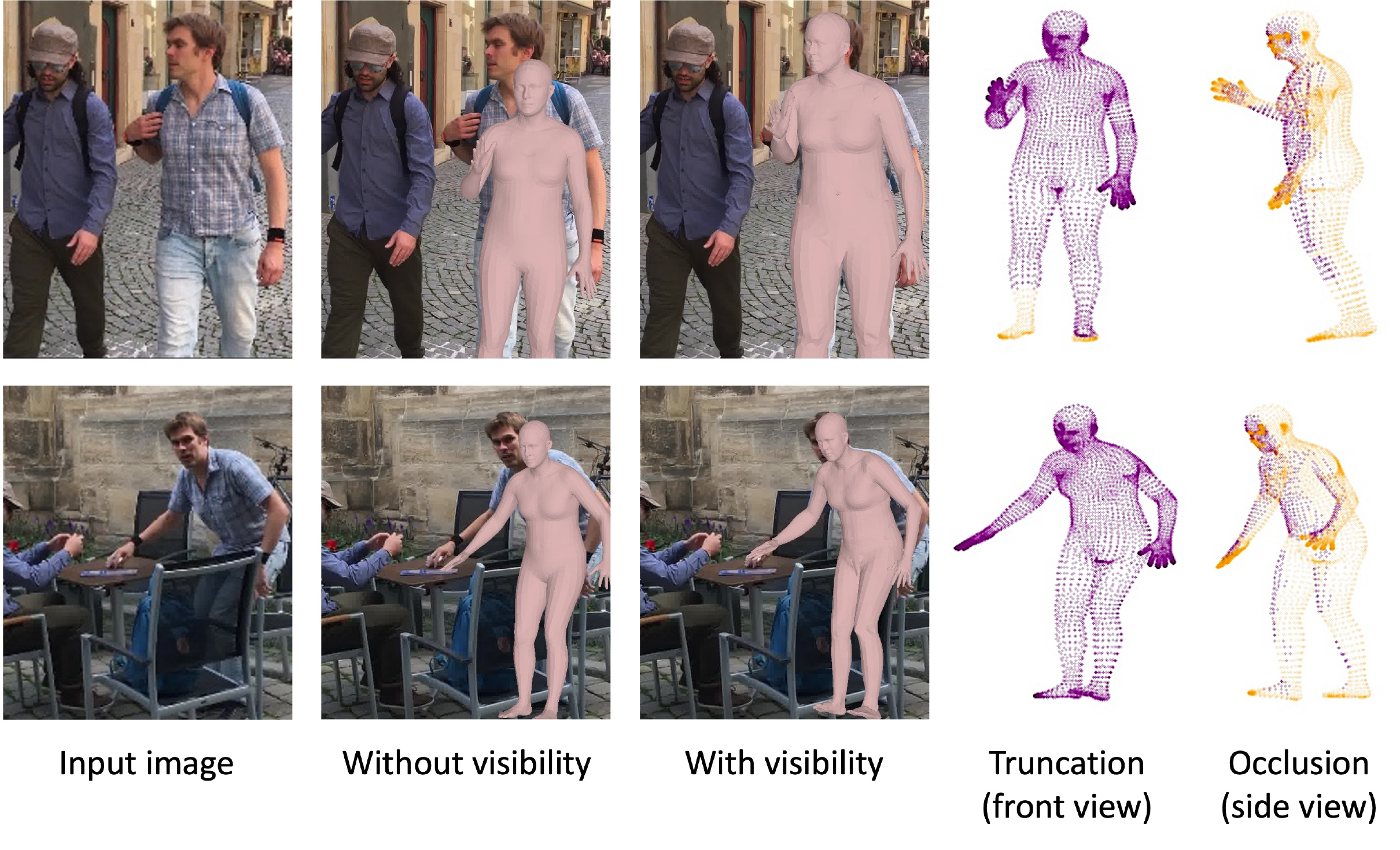}
    \end{center}
    \caption{\textbf{Dense human body estimation with/without visibility modeling.} We propose to learn dense visibility to improve human body estimation in terms of faithfulness to the input image and robustness to truncation (top) or occlusions (bottom). We show the estimated meshes without/with visibility modeling in columns 2-3 and the vertex visibility labels in columns 4-5 (\textcolor{purple}{purple:visible}, \textcolor{orange}{orange:invisible}).}
\label{fig:intro}
\end{figure}

%
Estimating 3D human pose and shape from monocular images is a crucial task for various applications such as performance retargeting, virtual avatars, and human action recognition.
It is a fundamentally challenging problem due to the depth ambiguity and the complex nature of human appearances that vary with articulation, clothing, lighting, viewpoint, and occlusions.
To represent the complicated 3D human bodies via compact parameters, model-based methods like SMPL~\cite{loper2015smpl} have been widely used in the community.
However, SMPL parameters represent human bodies in a holistic manner, causing their limited flexibility to fit real-world images faithfully via direct regression.
More importantly, the regression-based methods tend to fail when a human body is not fully visible in the image, \eg, occluded or out of frame~\cite{kocabas2021pare}.
In this work, we aim to learn human body estimation that is faithful to the input images and robust to partial-body cases.

%
Instead of directly regressing SMPL parameters, we train a neural network to predict the coordinate heatmaps in three dimensions for each human joint and mesh vertex.
The dense heatmap-based representation can preserve the spatial relationship in the image domain and model the uncertainty of predictions.
It is shown to be effective in localizing visible joints/vertices and flexible to fit an input image faithfully~\cite{sun2018integral,moon2018v2v,moon2019camera,moon2020i2l}.
Nonetheless, the x and y-axis heatmaps are defined in the image coordinates, which cannot represent the out-of-frame (\ie, truncated by image boundaries) body parts.
In addition, occlusions by objects or the human body itself could cause ambiguity for depth-axis predictions.
Without knowing which joints/vertices are visible, the network tends to produce erroneous outputs on partial-body images.
To address this, we propose \emph{Visibility-aware Dense Body (VisDB)}, a heatmap-based dense representation augmented by visibility.
Specifically, we train a network to predict binary truncation and occlusion labels along with the heatmaps for each human joint and vertex.
With the visibility modeling, the proposed network can learn to make more accurate predictions based on the observable cues.
%
%
In addition, the vertex-level occlusion predictions can serve as a depth ordering signal to constrain depth predictions.
Finally, by using visibility as the confidence of 3D mesh prediction, we demonstrate that VisDB is a powerful intermediate representation which allows us to regress and/or optimize SMPL parameters more effectively.
%
%
%
In Figure~\ref{fig:intro}, we show examples of truncation and occlusions as well as the dense human body estimations with and without visibility modeling.

%
Considering that most existing 3D human datasets lack dense visibility annotations, we obtain pseudo ground-truths from dense UV estimations~\cite{guler2018densepose}.
Given the estimated UV map of an image, we calculate the pixel-to-vertex correspondence by minimizing the distance of their UV coordinates.
Each vertex mapped to a human pixel is considered visible, and vice versa.
Note that this covers the cases of truncation, self-occlusions, and occlusions by other objects.
We further show that the dense vertex-to-pixel correspondence provides a good supervisory signal to localize vertices in the image space. 
Since dense UV estimations are based on part-wise segmentation masks which are robust to partial-body images, the dense correspondence loss can mitigate the inaccurate pseudo ground-truth meshes and better align the outputs with human silhouettes.
%
%
%
%
To demonstrate the effectiveness of our method, we conduct extensive experiments on multiple human datasets used by prior arts.
Both qualitative and quantitative results on the Human3.6M~\cite{ionescu2013human3}, 3DPW~\cite{von2018recovering}, 3DPW-OCC~\cite{von2018recovering,zhang2020object}, and 3DOH~\cite{zhang2020object} datasets show that learning visibility significantly improves the accuracy of dense human body estimation, especially on images with truncated or occluded human bodies.

The main contributions of our work are:
\begin{itemize}
    \item We propose VisDB, a heatmap-based human body representation augmented with dense visibility. 
    We train a neural network to predict the 3D coordinates of human joints and vertices as well as their truncation and occlusion labels. 
    We obtain pseudo ground-truths of visibility labels from image-based dense UV estimates, which are also used as additional supervision signal to better align our predictions with the input image.
    \item We show how the dense visibility predictions can be used for robust human body estimation. First, we exploit occlusion labels to supervise vertex depth predictions. Second, we regress and optimize SMPL parameters to fit VisDB (partial-body) outputs by using visibility as confidence weighting.
    %
    %
\end{itemize}

\section{Related Work}

\noindent {\bf Model-based human body estimation.}
Most existing methods on human body estimation adopt a model-based representation.
For instance, SMPL~\cite{loper2015smpl} is a widely-used statistical human body model that maps a set of pose $\theta \in \mathbb{R}^{72}$ and shape $\beta \in \mathbb{R}^{10}$ parameters to a 3D human mesh $V \in \mathbb{R}^{6890\times3}$.
In SMPL, $\theta$ represents the axis-angle 3D rotations of 24 joints, and $\beta$ is the top-10 PCA coefficients of a statistical human shape space.
Early methods iteratively optimize the SMPL parameters to fit the estimated 2D keypoints~\cite{bogo2016keep} or silhouettes~\cite{lassner2017unite}.
Several recent works~\cite{kanazawa2018end,pavlakos2018learning,omran2018neural,pavlakos2019texturepose,kolotouros2019learning,kolotouros2021prohmr} train a deep neural network to directly regress SMPL parameters from an input image.
However, the SMPL representation is not always informative enough for a network to learn as it embeds the articulated body shapes in a low dimensional space.
The regression-based methods often fail on truncation and occlusion cases since the networks tend to make holistic predictions based on certain body parts only~\cite{kocabas2021pare}.
Instead, we show that localizing 3D vertices is a more suitable task to learn for such scenarios.
The network needs to learn the relationship between the parameters and the shape as well in order to estimate accurate SMPL parameters.
%
%
%

\noindent {\bf Dense human body representations.}
To fit the complicated shapes more faithfully, dense human body representations have been proposed, including volumetric space~\cite{varol2018bodynet}, occupancy field~\cite{saito2019pifu,saito2020pifuhd}, dense UV correspondence~\cite{alldieck2019tex2shape,zeng20203d}, and 3D mesh~\cite{kolotouros2019convolutional,choi2020pose2mesh,moon2020i2l,lin2021end,lin2021mesh}.
Among these methods, I2L-MeshNet~\cite{moon2020i2l} proposes an efficient heatmap representation to estimate human joints and vertices in the image space and root-relative depth axis.
It can fit the input images accurately since heatmaps preserve the spatial relationship in image features extracted by a convolutional neural network (CNN).
Nonetheless, even when certain body parts are not visible in the image, this model is designed to localize all the joints and vertices within the image frame.
We show that it can negatively affect the model performance and emphasize the importance of additional visibility information.

\noindent {\bf Occlusion-aware methods.}
Several methods have been proposed to deal with the challenging scenarios where human bodies are partially truncated or occluded.
Muller~\etal~\cite{muller2021self} and Hassan~\etal~\cite{hassan2019resolving} introduce explicit modeling of human body self-contact and human-scene interactions, respectively.
These methods require ground-truth annotations which are hard to obtain.
Other methods leverage human-centric heatmaps, part segmentation masks, or dense UV estimations~\cite{guler2018densepose}, to increase the model robustness on truncated images~\cite{rockwell2020full}, crowded scenes (occluded by other people)~\cite{sun2021monocular} or general occlusions~\cite{xu2019denserac,guler2019holopose,zhang2020object,kocabas2021pare}.
Although effective in particular scenarios, most of them directly regress SMPL parameters which still suffer from the limited representation strength.
To the best our our knowledge, the proposed VisDB representation is the first to explicitly model dense human body visibility (including truncation and all occlusion scenarios), which is trained with pseudo ground-truth visibility labels from dense UV estimates.

\begin{figure*}[!t]
    \begin{center}
    \includegraphics[width=.9\linewidth]{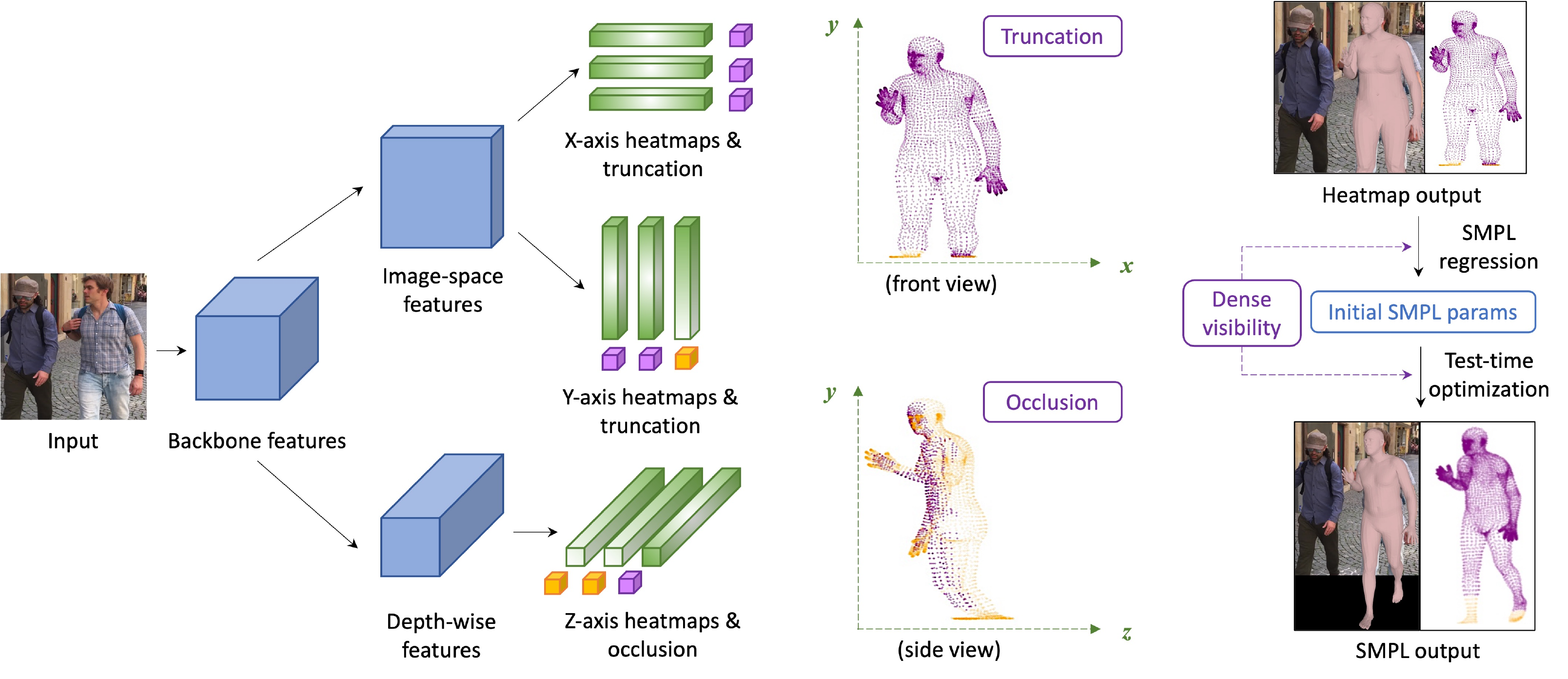}
    \end{center}
    \caption{\textbf{VisDB framework overview} (best viewed in color).
    Given an input image, the network extracts features in the image and depth coordinates, from where we predict the x, y, z heatmaps for each human joint and vertex.
    In addition, we predict a binary visibility label (\textcolor{purple}{purple:visible}, \textcolor{orange}{orange:invisible}) of each axis, \ie, x-truncation, y-truncation,
    To obtain a more regularized and complete human body, we train a regression network to estimate SMPL parameters based on the dense 3D coordinates and visibility.
    At test time, we can further optimize the regressed SMPL parameters to fit the partial-body predictions from heatmaps. 
    }
\label{fig:framework}
\end{figure*}

\section{Approach}

%
We illustrate our overall framework in Figure~\ref{fig:framework}.
In Section~\ref{sec:prelim}, we describe a heatmap-based representation which we build our method upon.
Then, we introduce the proposed Visibility-aware Dense Body (VisDB) in Section~\ref{sec:visdb}.
Each human joint and mesh vertex is represented by 1) three 1D heatmaps (x, y, z dimensions) which define its 3D coordinate and 2) three binary labels indicating its visibility in three dimensions.
We train a network model to predict the dense heatmaps and visibility, which represents a partial body faithful to the input image.
The visibility estimations can be interpreted as depth ordering signals or prediction confidence.
In Section~\ref{sec:depth}, we design a visibility-guided depth ordering loss to self-supervise depth estimation.
In Section~\ref{sec:smpl}, we show that VisDB outputs can be used to fit SMPL models accurately and efficiently.
We train a regression network to estimate SMPL parameters based on the joint and vertex coordinates as well as their visibility labels.
During inference, we initialize the SMPL parameters by the regressor and further optimize them to align with the VisDB predictions.
Finally, in Section~\ref{sec:dense-uv}, we exploit dense UV correspondence to obtain robust pseudo labels of visibility and weakly supervise vertex localization in the image space.

%
\subsection{Preliminaries: Heatmap-based Representation}
\label{sec:prelim}
Given an input image, a prior heatmap-based method~\cite{moon2020i2l} estimates three 1D heatmaps $H=\{H^x, H^y, H^z\}$ for each human joint and mesh vertex.
The x and y-axis heatmaps $H^x, H^y$ are defined in the image space, and the z-axis heatmaps $H^z$ are defined in the depth space relative to root joint.
We denote the joint heatmaps as $H_J \in \mathbb{R}^{N_J \times D \times 3}$ and vertex heatmaps as $H_V \in \mathbb{R}^{N_V \times D \times 3}$, where $N_J$ is the number of joints, $N_V$ is the number of vertices, and $D$ is the heatmap resolution.
The heatmaps are predicted based on image features $F \in \mathbb{R}^{c \times h \times w}$ extracted by a backbone network as follows:
\begin{equation}
\begin{split}
     H^x &= f^{\text{ 1D},x} (\text{avg}^y (f^\text{ up} (F))), \\
     H^y &= f^{\text{ 1D},y} (\text{avg}^x (f^\text{ up} (F))),  \\
     H^z &= f^{\text{ 1D},z} (\psi (\text{avg}^{x,y} (F)))),
\end{split}
\label{eq:heatmaps}
\end{equation}
%
where $f^{\text{1D},i}$ is 1-by-1 1D convolution for the $i$-th axis heatmaps, $\text{avg}^i$ is i-axis marginalization by averaging, $f^\text{ up}$ denotes up-sampling by deconvolution, and $\psi$ is a 1D convolution layer followed by reshaping operation.
Finally, the continuous 3D coordinates of joints $J \in \mathbb{R}^{N_J \times 3}$ and vertices $V \in \mathbb{R}^{N_V \times 3}$ can be obtained by applying soft-argmax on the discrete heatmaps $H_J$ and $H_V$, respectively.
More details can be found in~\cite{moon2020i2l} and supplementary material.

\subsection{Visibility-aware Dense Body}
\label{sec:visdb}
Heatmap-based representations are shown effective in estimating human bodies in the image space.
However, they often fail when the human bodies are occluded or truncated since the predictions are based on spatial image features and limited by the image boundaries.
Without knowing which joints/vertices are invisible, fitting a SMPL model on the entire body tends to generate erroneous outputs.
To deal with more practical scenarios where only partial bodies are visible, we make the following adaptations to a heatmap-based representation:
1) To augment the x and y-axis heatmaps, we predict binary truncation labels $S^x,S^y$, indicating whether a joint or vertex is within the image frame,
2) For the z-axis heatmaps, we predict a binary occlusion label $S^z$ which specifies the depth-wise visibility.
The visibility labels are predicted in a similar fashion as the heatmaps in Eq.~\eqref{eq:heatmaps}:
\begin{equation}
\begin{split}
     S^x &= \sigma (\text{avg}^x (g^{\text{1D},x} (\text{avg}^y (f^\text{ up} (F))))),\\
     S^y &= \sigma (\text{avg}^y (g^{\text{1D},y} (\text{avg}^x (f^\text{ up} (F))))),\\
     S^z &= \sigma (\text{avg}^z (g^{\text{1D},z} (\psi (\text{avg}^{x,y} (F))))),
\label{eq:visibility}
\end{split}
\end{equation}
%
where $g^{\text{1D}}$ is a 1-by-1 1D convolutional layer similar to $f^{\text{1D}}$ and $\sigma$ is a sigmoid operator.
We then concatenate the $\{S^x,S^y,S^z\}$ predictions to obtain joint visibility $S_J \in \mathbb{R}^{N_J \times 3}$ and vertex visibility $S_V \in \mathbb{R}^{N_V \times 3}$.
By applying the soft-argmax operators to the predicted 1D heatmaps, the final output of our network becomes $\{J, V, S_J, S_V\}$, referred to as Visibility-aware Dense Body (VisDB).
With the visibility information, the network model can learn to focus on the visible body parts and push the invisible parts towards the image boundaries.
In our experiments (Table~\ref{tab:ablation_heatmap}), we demonstrate that visibility modeling significantly reduces the errors of visible vertices.
Moreover, the visibility labels can be seen as the confidence of coordinate predictions, which are essential to mesh regularization and completion via SMPL model fitting as described in Section~\ref{sec:smpl}.

We denote the ground-truth VisDB as $\{J^*, V^*, {S_J}^*, {S_V}^*\}$ and train the network by using the following losses.
%
The joint coordinate loss $\mathcal{L}_{joint}$ is defined as:
\begin{equation}
    \mathcal{L}_{joint} = \|J - J^*\|_1 .
\label{eqn:joint}
\end{equation}
%
%
The vertex coordinate loss $\mathcal{L}_{vert}$ is defined as:
\begin{equation}
    \mathcal{L}_{vert} = \|V - V^*\|_1 .
\label{eqn:vertex}
\end{equation}
%
%
We also regress the joints from vertices using a pre-defined regressor $W \in \mathbb{R}^{N_V \times N_J}$ and calculate a regressed-joint loss $\mathcal{L}_{r-joint}$:
\begin{equation}
    \mathcal{L}_{r-joint} = \|WV - J^*\|_1 .
\label{eqn:r-joint}
\end{equation}
%
%
Similar to~\cite{moon2020i2l}, we apply losses on the mesh surface normal and edge length as shape regularization.
The normal loss $\mathcal{L}_{norm}$ and edge loss $\mathcal{L}_{edge}$ are:
\begin{align}
    \mathcal{L}_{norm} &= \sum_f \sum_{\{v_i,v_j\} \subset f} 
        \Big| \Big\langle \frac{v_i - v_j}{\| v_i - v_j \|_2}, {n_f}^* \Big\rangle \Big| , \\
    \mathcal{L}_{edge} &= \sum_f \sum_{\{v_i,v_j\} \subset f} 
                          \Big| \|v_i - v_j\|_2 - \|{v_i}^* - {v_j}^*\|_2 \Big| ,
\end{align}
where $f$ is a mesh surface, $n_f$ is the unit normal vector of f, and $v_i,v_j$ are the coordinates of vertex $i$ and $j$, respectively.
%
%
Finally, we define the joint and vertex visibility loss $\mathcal{L}_{vis}$ with binary cross entropy (BCE):
\begin{equation}
    \mathcal{L}_{vis} = \text{BCE}(S_J, {S_J}^*) + \text{BCE}(S_V, {S_V}^*) .
\end{equation}
The VisDB prediction is illustrated in Figure~\ref{fig:framework} (left).

%
\subsection{Resolving Depth Ambiguity via Visibility}
\label{sec:depth}
Vertex-level visibility can not only be seen as model confidence for SMPL fitting but also provide depth ordering information.
Intuitively, visible vertices should have lower depth value compared to the invisible vertices projected to the same pixel.
We observe that VisDB network generally predicts accurate 2D coordinates and visibility, but sometimes fails at depth predictions when the human body occludes itself and the pose is less common in the training datasets.
To resolve the depth ambiguity in self-occlusion cases, we propose a depth ordering loss $\mathcal{L}_{depth}$ based on vertex visibility as follows:
\begin{equation}
    \mathcal{L}_{depth} = \sum_x \sum_y \text{ReLU} \Big( \max_{v \in Q(x,y)} v^z - \min_{\overline{v} \in \overline{Q}(x,y)} \overline{v}^z \Big) ,
\end{equation}
where $Q(x,y)$ is the set of vertices projected to a discretized image coordinate $(x,y)$ which belong to the front (occluding) part, and $\overline{Q}$ contains the vertices of the back (occluded) part(s).
The definition can be written as:
\begin{equation}
\begin{split}
    Q(x,y) &= \Big\{ v \big| v\mapsto(x,y) \wedge P(v) = p^*(x,y) \Big\} 
    \\
    \overline{Q}(x,y) &= \Big\{ v \big| v\mapsto(x,y) \wedge P(v) \neq p^*(x,y) \Big\} ,
\end{split}
\end{equation}
where $\mapsto$ denotes the discrete projection and $P(v)$ is the part label of vertex $v$ defined in DensePose~\cite{guler2018densepose}. 
We define the front part $p^*(x,y)$ by finding the vertex with highest z-axis visibility score $s^z$ as:
\begin{equation}
    p^*(x,y) = P \Big(\argmax_{v\mapsto(x,y)} {s_v}^z \Big) .
\end{equation}
$\mathcal{L}_{depth}$ is designed to push the self-occluded part(s) $\overline{Q}$ to the back and non-occluded part $Q$ to the front, where the occlusion information is given by the z-axis visibility.
Note that we compare the maximum depth (back side) of $Q$ and the minimum depth (front side) of $\overline{Q}$, and thus $\mathcal{L}_{depth}$ will be nonzero if the depth ordering disagrees with occlusion prediction and zero if the parts do not overlap anymore.
Since this loss depends on accurate visibility estimations, we only apply it during the fine-tuning stage.

\subsection{SMPL Fitting from Visible Dense Body}
\label{sec:smpl}
%
From the VisDB predictions, we can obtain the 3D coordinates and visibility of human joints and vertices.
While the partial-body outputs are faithful to the input image from the front view, they sometimes look abnormal from a side view or contain rough surfaces. 
To regularize the body shape and complete the truncated parts, we perform model fitting on the visible dense body predictions.
%
%
Given the coordinates and visibility of joints and vertices, we train a regression network to estimate SMPL pose $\theta \in \mathbb{R}^{72}$ and shape $\beta \in \mathbb{R}^{10}$ parameters.
The regressed parameters are then forwarded to the SMPL model to generate the mesh coordinates denoted as $\text{SMPL}(\theta,\beta) \in \mathbb{R}^{N_V \times 3}$.
Unlike prior art~\cite{moon2020i2l} which regresses a SMPL model from all the joints regardless of their visibility, our VisDB representation allows us to fit the visible partial body only.
The training objectives of the SMPL regressor include SMPL parameter error, vertex error, joint error, and the negative log-likelihood of a pose prior distribution.
The SMPL parameter loss $\mathcal{L}_{smpl}$ is defined as:
\begin{equation}
    \mathcal{L}_{smpl} = \|\theta - \theta^*\|_1 + \|\beta - \beta^*\|_1 ,
\label{eqn:smpl}
\end{equation}
where $\theta^*$ and $\beta^*$ are the ground-truth pose and shape parameters.
The SMPL vertex loss $\mathcal{L}_{smpl-vert}$ and joint loss $\mathcal{L}_{smpl-joint}$ are defined similarly as in Eq.~\eqref{eqn:vertex} and ~\eqref{eqn:r-joint} but weighted by visibility $S_V,S_J$ as:
\begin{equation}
    \mathcal{L}_{smpl-vert} = S_V \odot \|\text{SMPL}(\theta,\beta) - {V_c}^*\|_1 ,
\label{eqn:smpl-vert}
\end{equation}
\begin{equation}
    \mathcal{L}_{smpl-joint} = S_J \odot \| W \text{SMPL}(\theta,\beta) - {J_c}^*\|_1 ,
\label{eqn:smpl-joint}
\end{equation}
where $\odot$ denotes element-wise multiplication, and $({V_c}^*,{J_c}^*)$ are the ground-truth root-relative coordinates of vertices and joints in the camera space.
Ideally, the VisDB network makes more confident predictions on the clearly visible joints and vertices.
Hence, we see the visibility labels as prediction confidence and use them to weight the coordinate losses.
In addition, we apply a pose prior loss $\mathcal{L}_{prior}$ using a fitted Gaussian Mixture Model (GMM) provided by~\cite{pavlakos2019expressive}:
\begin{equation}
    \mathcal{L}_{prior} = - \text{log} \Big( \sum_i G_i(\theta) \Big) ,
\label{eqn:prior}
\end{equation}
where $G_i$ is the $i$-th component of GMM.

We observe that the regressed SMPL meshes roughly capture the human pose and shape but do not always align with the VisDB predictions in details.
Therefore, we use the regressed parameters as initialization and propose efficient test-time optimization to further optimize the SMPL parameters against VisDB predictions.
For this optimization, we apply similar losses as in Eq.~\eqref{eqn:smpl-vert}-\eqref{eqn:prior}, except that the ground-truths $\{{V_c}^*,{J_c}^*\}$ are replaced by the VisDB predictions converted into root-relative coordinates in the camera space.
Please refer to the supplemental material for details on estimating the root joint coordinate in the camera space.
%
%
Since we initialize the SMPL parameters by the regression network and the use strong supervisory signal, \ie, 3D joint and vertex coordinates, the test-time optimization only takes around 100 iterations to converge using an Adam optimizer~\cite{kingma2014adam}.
We illustrate the process of SMPL regression and optimization in Figure~\ref{fig:framework} (right).

\begin{figure}[!t]
    \begin{center}
    \includegraphics[width=0.6\linewidth]{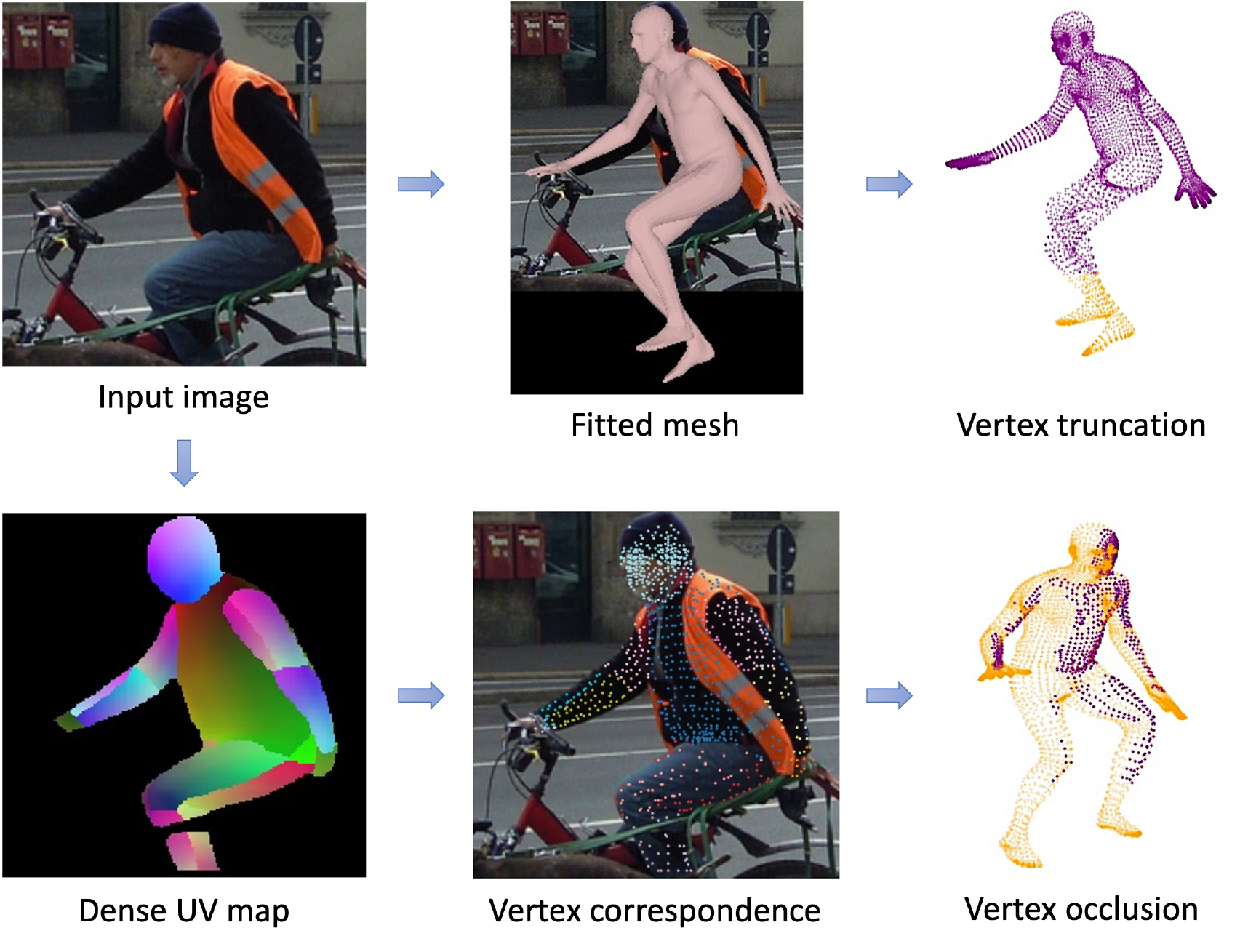}
    \end{center}
    \caption{\textbf{Dense UV correspondence and visibility labels.} 
    Given an input image, we obtain a fitted SMPL mesh and dense UV estimation from off-the-shelf algorithms.
    To acquire the dense visibility labels for training, we identify the truncated vertices from the fitted mesh.
    From the dense UV map, we calculate the pixel-to-vertex correspondence to obtain pseudo ground-truths of vertex occlusions as well as image-space coordinates for weak supervision.
    }
\label{fig:dense_uv}
\end{figure}

\subsection{Exploiting Dense UV Correspondence}
\label{sec:dense-uv}

Most existing 3D human datasets do not provide joint visibility labels, and none annotates vertex visibility.
To train our VisDB network, we obtain pseudo ground-truths from the fitted SMPL meshes and dense UV estimations.
For x and y-axis truncation, we can simply identify the truncated joints/vertices by projecting the fitted mesh onto the image plane.
Occlusion, however, cannot be easily inferred from the input image or fitted mesh alone.
One can estimate self-occlusion by rendering a fitted mesh, but this does not capture occlusions by other objects.
More importantly, the fitting algorithm used to get the pseudo ground-truth meshes is not robust to partial-body cases.
To address this, we propose to exploit dense UV correspondence between the input image and a SMPL mesh.
Dense UV estimation provides the part-based segmentation mask of a human body as well as continuous UV coordinates of each human pixel, which are robust to truncation and occlusions.
We calculate the UV coordinate of each pixel by applying an off-the-shelf dense UV estimation method~\cite{guler2018densepose}.
For each human pixel $p$, we then find the corresponding mesh vertex $v$ whose UV coordinate is closest to the pixel.
The pixel-to-vertex $M_P$ and vertex-to-pixel $M_V$ mappings can be expressed as:
\begin{equation}
\begin{split}
    M_P &= \big\{ p \rightarrow v \big| v = \text{argmin}_{v'} \big\| \text{UV}(v') - \text{UV}(p) \big\|_2 \forall p \big\} \\
    M_V &= \big\{ v \rightarrow \{p'\} \big| M_P(p') = v \forall v \big\}.
\end{split}
\label{eqn:mappings}
\end{equation}
A vertex mapped to at least one pixel is labeled as visible or occluded otherwise.
%

Similar to~\cite{guler2019holopose,xu2019denserac,zeng20203d}, we also utilize the dense vertex-pixel correspondence as weak supervision for better alignment with the human silhouettes. 
For each vertex $v$, we calculate the center of its corresponding pixels $M_V(v)$ and define a UV correspondence loss $\mathcal{L}_{uv}$ as:
\begin{equation}
    \mathcal{L}_{uv} = 
    \sum_{v} {s_v}^z \Big\|v^{x,y} - 
    \sum_{p \in M_V(v)} \frac{p}{|M_V(v)|} \Big\|_1 ,
\label{eqn:uv}
\end{equation}
where $v^{x,y}$ is the 2D projection of vertex $v$ and ${s_v}^z$ is the binary occlusion label with ${s_v}^z=1$ indicating that the vertex $v$ is visible.
The UV correspondence loss can not only mitigate the inaccurate pseudo ground-truth meshes, but improve the faithfulness to human silhouettes since it is based on segmentation mask predictions.
We empirically discover that this direct vertex-level supervision is more efficient and effective for VisDB training compared to rendering-based losses~\cite{xu2019denserac,dwivedi2021learning}.
The proposed vertex-pixel correspondence and visibility labeling are illustrated in Figure~\ref{fig:dense_uv}.

\subsection{Model Training and Inference}
\label{sec:details}
We first train the VisDB network on 3D data with mesh annotations, then fine-tune it on all training data by adding the depth ordering and UV correspondence losses.
The regressor network is trained to estimate the SMPL parameters based on the estimated coordinates and visibility of joints and vertices.
During inference, we apply optional optimization on the regressed SMPL parameters to best align with the VisDB predicted mesh.
For the VisDB network backbone, we use a ResNet50~\cite{he2016deep} model pre-trained on the ImageNet dataset~\cite{deng2009imagenet}.
%
%
The weights are updated by the Adam optimizer~\cite{kingma2014adam} with a mini-batch size of 64.
We represent a human body by $N_J=30$ joints and $N_V=6890$ vertices, and the heatmap resolution $D=64$.
In addition, we use the ground-truth bounding boxes to crop the human region from an input image and resize it to 256$\times$256. 
The bounding boxes of testing data are estimated by a pre-trained Mask R-CNN~\cite{he2017mask} model if not available in the dataset.
We apply common data augmentations such as random scaling (±25\%), rotation (±45°), horizontal flip, and color jittering (±20\%) during training.
Considering that truncation and occlusion examples are rare in most 3D human datasets, we include random occlusion masks and bounding box shifting (±25\%) as additional augmentations to increase the partial-body/whole-body ratio.
%
%
Our models are implemented with PyTorch~\cite{paszke2019pytorch} and trained with NVIDIA Tesla V100 GPUs.
%
%
More implementation details are presented in the supplemental material.

\section{Experiments}

\subsection{Datasets and Metrics}

Following most prior arts, we adopt mixed 2D-3D training on the MSCOCO~\cite{lin2014microsoft}, Human3.6M~\cite{ionescu2013human3}, MuCo-3DHP~\cite{mehta2018single}, and 3DPW~\cite{von2018recovering} datasets.
The pseudo ground-truth meshes of Human3.6M and MSCOCO are obtained by applying SMPLify-X~\cite{pavlakos2019expressive} to fit the joint annotations.
We evaluate our models on the Human3.6M, 3DPW, 3DPW-OCC~\cite{von2018recovering,zhang2020object}, and 3DOH~\cite{zhang2020object} testing sets.
Note that 3DOH is composed of images with object occlusions and 3DPW-OCC contains a subset of 3DPW sequences where the human bodies are partially occluded.
%
%
%
%
%
%
%
%
For quantitative evaluation, we calculate the common joint and vertex error metrics in the camera space and report them in millimeters (mm), including MPJPE (mean per-joint position error)~\cite{ionescu2013human3}, PA-MPJPE (Procrustes-aligned mean per-joint position error)~\cite{zhou2018monocap}, and MPVE (mean per-vertex error)~\cite{pavlakos2018learning}.

\begin{table*}[!t]
\scriptsize
\centering
\caption{\textbf{Quantitative evaluations on Human3.6M~\cite{ionescu2013human3} and 3DPW~\cite{von2018recovering}.} To align the settings, we train our baseline, I2L-MeshNet~\cite{moon2020i2l}, on the same datasets, and denote it by I2L-MeshNet$^\dagger$. Both our mesh and SMPL parameter outputs perform favorably against the prior state-of-the-arts.}
%
\setlength\tabcolsep{4pt}
\begin{tabular}{lcccccc}
    \toprule
    & & \multicolumn{2}{c}{Human3.6M}  & \multicolumn{3}{c}{3DPW} \\
    \cmidrule(lr){3-4}\cmidrule(lr){5-7}
    Method & Output & MPJPE$\downarrow$ & PA-MPJPE$\downarrow$ & MPJPE$\downarrow$ & PA-MPJPE$\downarrow$ &  MPVE$\downarrow$ \\
    \midrule
    GraphCMR~\cite{kolotouros2019convolutional} & Mesh & -    &  50.1 &  -    & 70.2 & -   \\
    Pose2Mesh~\cite{choi2020pose2mesh}          & Mesh & 64.9 &  47.0 &  89.2 & 58.9 & 109.3  \\
    I2L-MeshNet~\cite{moon2020i2l}              & Mesh & 55.7 &  41.1 &  93.2 & 57.7 & 109.2 \\
    I2L-MeshNet$^\dagger$~\cite{moon2020i2l}    & Mesh & - & -    & 84.5 & 51.1 & 98.2 \\
    METRO~\cite{lin2021end}                     & Mesh & 54.0 &  36.7 &  77.1 & 47.9 & 88.2 \\
    Mesh Graphormer~\cite{lin2021mesh}          & Mesh & 51.2 &  {\bf34.5} &  74.7 & 45.6 & 87.7 \\
    VisDB  (mesh)                               & Mesh & {\bf51.0} &  {\bf34.5} & {\bf73.5} & {\bf44.9} & {\bf85.5} \\
    \midrule
    NBF~\cite{omran2018neural}                  & Param & -    & 59.9     & -  & - & - \\
    HMR~\cite{kanazawa2018end}                  & Param & 88.0 &  56.8 &  -    & 81.3 & -   \\
    DenseRaC~\cite{xu2019denserac}              & Param & 76.8 &  48.0    &  - & - & - \\
    I2L-MeshNet~\cite{moon2020i2l}              & Param & -    &  -    &  100.0 & 60.0 & 121.5 \\
    OOH~\cite{zhang2020object}                  & Param & -    &  41.7    &  - & - & - \\
    SPIN~\cite{kolotouros2019learning}          & Param & -    &  41.1 &  -    & 59.2 & 116.4  \\
    I2L-MeshNet$^\dagger$~\cite{moon2020i2l}    & Param & - & -   & 88.0  & 55.5 & 102.3 \\
    DSR~\cite{dwivedi2021learning}              & Param & 60.9 &  40.3    & 85.7 & 51.7 & 99.5 \\
    VIBE~\cite{kocabas2020vibe}                 & Param & 65.6 & 41.4 & 82.0 & 51.9 & 99.1 \\
    TCMR~\cite{choi2021beyond}                  & Param & 62.3 & 41.1 & - & - & - \\
    DecoMR~\cite{zeng20203d}                    & Param & 60.6 & 39.3 & - & - & - \\
    PARE~\cite{kocabas2021pare}                 & Param & - & - & 79.1 & 46.4 & 94.2 \\
    VisDB (param)                               & Param & {\bf50.0} & {\bf33.8} & {\bf72.1} & {\bf44.1} & {\bf83.5} \\
    \bottomrule
\end{tabular}
\label{tab:sota}
\end{table*}

\subsection{Quantitative Comparisons}
%
\noindent {\bf Human3.6M and 3DPW.}
In Table~\ref{tab:sota}, we compare the performance of our method and prior arts on the Human3.6M~\cite{ionescu2013human3} and 3DPW~\cite{von2018recovering} datasets.
For VisDB and I2L-MeshNet~\cite{moon2020i2l}, we report the results of both heatmap-based mesh outputs (mesh) and SMPL parameters (param).
%
%
Our SMPL parameters are obtained from regression and test-time optimization.
Note that each method uses different network backbone, human body representation, training datasets, and inference strategy.
For instance, METRO~\cite{lin2021end} and Mesh Graphormer~\cite{lin2021mesh} adopt a transformer-based~\cite{vaswani2017attention} network while the others use CNN backbones.
VIBE~\cite{kocabas2020vibe} and TCMR~\cite{choi2021beyond} are video-based approaches whereas the others only take images as input.
Despite these differences, VisDB performs favorably against prior methods in term of most evaluation metrics.
Particularly, our method achieves larger performance gains on the 3DPW dataset since it contains more truncation and occlusion cases.
The VisDB performance is most directly comparable with I2L-MeshNet~\cite{moon2020i2l} as we adopt similar training settings.
For fair comparisons, we re-train its model on the same datasets and denote it as I2L-MeshNet$^\dagger$.
The results demonstrate that our visibility learning improves both the mesh and SMPL outputs significantly.
In prior literature, SMPL parameters generally lead to higher errors compared to dense mesh outputs, which we conjecture is caused by the difficulty to directly regress low-dimensional parameters.
On the contrary, VisDB is a powerful intermediate representation that provides dense 3D information of visible partial body, allowing us to regress and optimize SMPL parameters more accurately.
%
%
%
In our experiments, we observe that VisDB (mesh) captures the human silhouettes better but VisDB (param) produces lower errors since the ground-truth meshes are also regularized by SMPL representation.

\begin{table*}[!t]
\scriptsize
\centering
\caption{\textbf{Quantitative evaluations on 3DOH~\cite{zhang2020object} and 3DPW-OCC~\cite{von2018recovering,zhang2020object}.} We compare VisDB with prior occlusion-aware methods to demonstrate its robustness on partial-body cases. For VisDB and I2L-MeshNet$^\dagger$~\cite{moon2020i2l}, We report both the mesh and SMPL parameter (mesh/param) results.}
%
\setlength\tabcolsep{6pt}
\begin{tabular}{lrrrrr}
    \toprule
    & \multicolumn{2}{c}{3DOH}  & \multicolumn{3}{c}{3DPW-OCC} \\
    \cmidrule(lr){2-3}\cmidrule(lr){4-6}
    Method & MPJPE$\downarrow$ & PA-MPJPE$\downarrow$ & MPJPE$\downarrow$ & PA-MPJPE$\downarrow$ &  MPVE$\downarrow$ \\
    \midrule
    OOH~\cite{zhang2020object} & - & 58.5 & - & 72.2 & - \\
    I2L-MeshNet$^\dagger$~\cite{moon2020i2l} & 67.0/69.3 & 46.3/47.9 & 96.5/98.0 & 61.0/62.6 & 120.2/127.0 \\
    PARE~\cite{kocabas2021pare} & 63.3 & 44.3 & 91.4 & 57.4 & 115.3 \\
    VisDB & {\bf62.1/60.9} & {\bf43.2/42.7} & {\bf90.3/87.3} & {\bf57.1/56.0} & {\bf114.0/110.5} \\
    \bottomrule
\end{tabular}
\label{tab:partial-body}
\end{table*}

\noindent {\bf 3DPW-OCC and 3DOH.}
To emphasize the robustness on partial-body images, we further evaluate on two occlusion datasets: 3DPW-OCC~\cite{von2018recovering,zhang2020object} and 3DOH~\cite{zhang2020object}.
As shown in Table~\ref{tab:partial-body}, VisDB produces lower errors on both datasets compared to prior occlusion-aware methods.
While I2L-MeshNet$^\dagger$ performs considerably worse on these images, the errors by our model remain relatively low.

\subsection{Ablation Studies}

\begin{table}[!t]
\scriptsize
\centering
\caption{\textbf{Ablation studies of VisDB.} We compare the joint/vertex errors of VisDB mesh outputs on 3DPW~\cite{von2018recovering} with/without individual components. The results show that truncation modeling (${\mathcal{L}_{vis}}^{x,y}$), occlusion modeling (${\mathcal{L}_{vis}}^{z}$), depth ordering loss $\mathcal{L}_{depth}$, and UV correspondence loss $\mathcal{L}_{uv}$ each reduces the errors by a clear margin.}
\setlength\tabcolsep{6pt}
\begin{tabular}{cccc ccc}
    \toprule
    ${\mathcal{L}_{vis}}^{x,y}$ & ${\mathcal{L}_{vis}}^{z}$ & $\mathcal{L}_{depth}$ & $\mathcal{L}_{uv}$ & MPJPE & PA-MPJPE & MPVE \\
    \midrule
           &        &        &        & 84.5 & 51.1 & 98.2 \\
           & \cmark & \cmark & \cmark & 79.4 & 47.8 & 91.1 \\
    \cmark &        & \cmark & \cmark & 75.8 & 45.5 & 88.0 \\
    \cmark & \cmark &        & \cmark & 77.3 & 46.3 & 88.9 \\
    \cmark & \cmark & \cmark &        & 74.9 & 45.6 & 87.1 \\
    \cmark & \cmark & \cmark & \cmark & {\bf73.5} & {\bf44.9} & {\bf85.5} \\
    \bottomrule
\end{tabular}
\label{tab:ablation_heatmap}
\end{table}

\noindent {\bf VisDB network training.}
To evaluate the contribution of individual components in our method, we perform ablation studies on the 3DPW dataset~\cite{von2018recovering}.
Table~\ref{tab:ablation_heatmap} shows the performance of VisDB mesh outputs with/without truncation modeling ${\mathcal{L}_{vis}}^{x,y}$, occlusion modeling ${\mathcal{L}_{vis}}^{z}$, depth ordering loss $\mathcal{L}_{depth}$, and dense UV correspondence loss $\mathcal{L}_{uv}$.
Without ${\mathcal{L}_{vis}}^{x,y}$, ${\mathcal{L}_{vis}}^{z}$, $\mathcal{L}_{depth}$, and $\mathcal{L}_{uv}$, the vertex error increases by 6.3mm, 3.1mm, 3.9mm, and 1.9mm, respectively.
These results show that both visibility modeling and depth ordering loss play a crucial role in VisDB training.

\noindent {\bf SMPL parameter fitting.}
In Table~\ref{tab:ablation_smpl}, we quantitatively compare the SMPL models obtained from different methods.
Given an estimated VisDB mesh, we can regress the SMPL parameters and/or optimize them during inference, and each process can be done with/without dense visibility weighting (Eq.~\eqref{eqn:smpl-vert} and~\eqref{eqn:smpl-joint}).
By using visibility, the mean vertex error of regressed SMPL models drops by 8.7mm.
With the proposed test-time optimization, we can further reduce the error by 3.8mm.

\begin{table}[!t]
\scriptsize
\centering
\caption{\textbf{Ablation studies of SMPL models}. We report the performance of SMPL outputs on the 3DPW dataset~\cite{von2018recovering}, which shows the effectiveness of our optimization and the importance of visibility in both regression and optimization work flows.}
%
\setlength\tabcolsep{6pt}
\begin{tabular}{llccc}
    \toprule
    Regression & Optimization & MPJPE & PA-MPJPE & MPVE \\
    \midrule
    -          & -            & 73.5 & 44.9 & 85.5 \\
    w/o vis    & -            & 79.0 & 48.8 & 96.2 \\
    w/o vis    & w/o vis      & 77.6 & 47.0 & 93.9 \\
    w/ vis     & -            & 74.9 & 45.3 & 87.3 \\
    w/ vis     & w/ vis       & {\bf72.1} & {\bf44.1} & {\bf83.5} \\
    \bottomrule
\end{tabular}
\label{tab:ablation_smpl}
\end{table}

\begin{figure*}[!t]
\scriptsize
\centering
\setlength\tabcolsep{1pt}
\begin{tabular}[t]{cccccc}
\includegraphics[height=20mm]{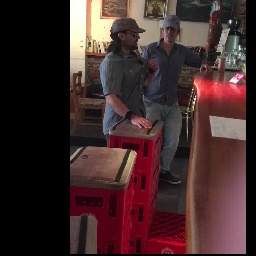} &
\includegraphics[height=20mm]{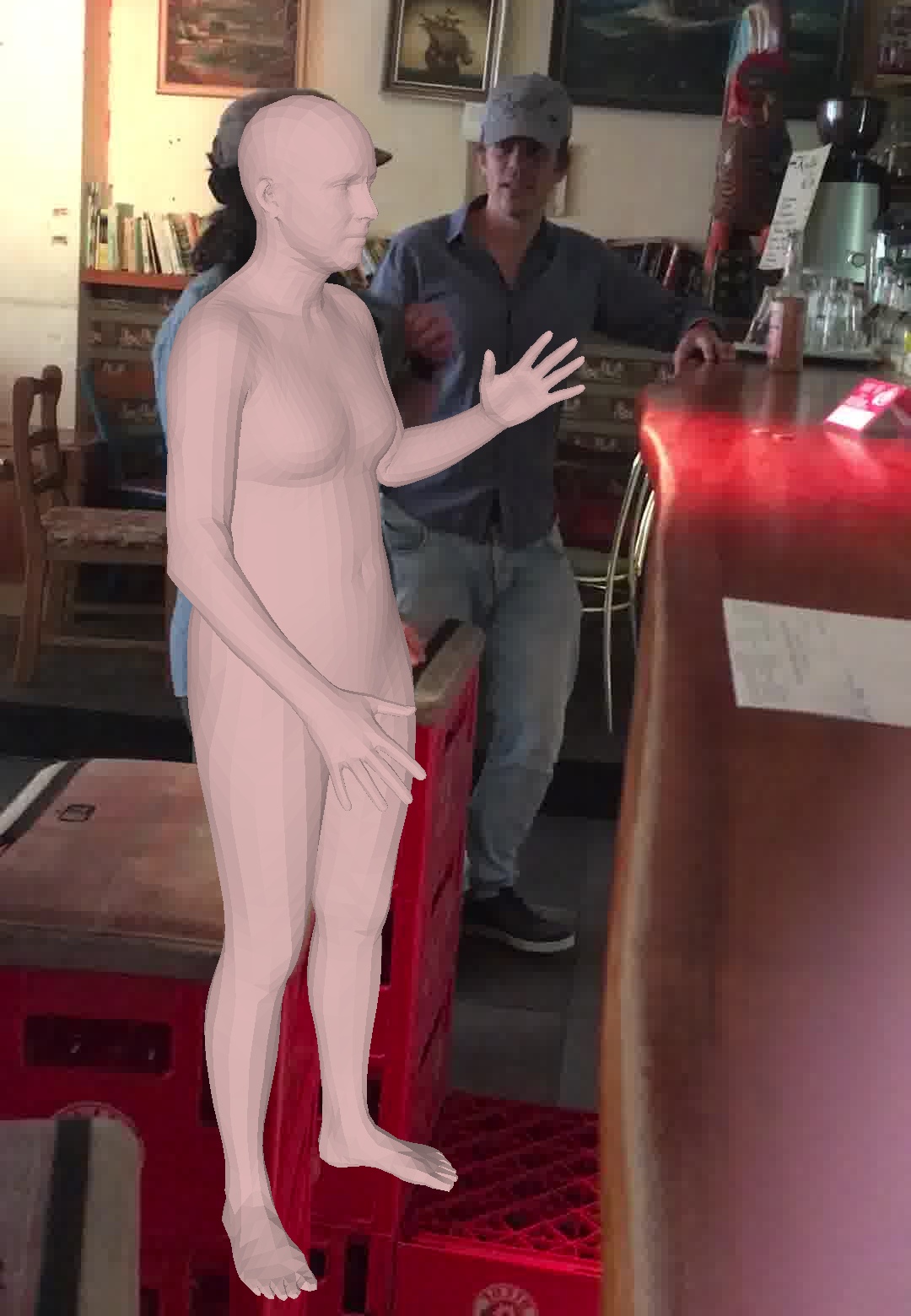} & 
\includegraphics[height=20mm]{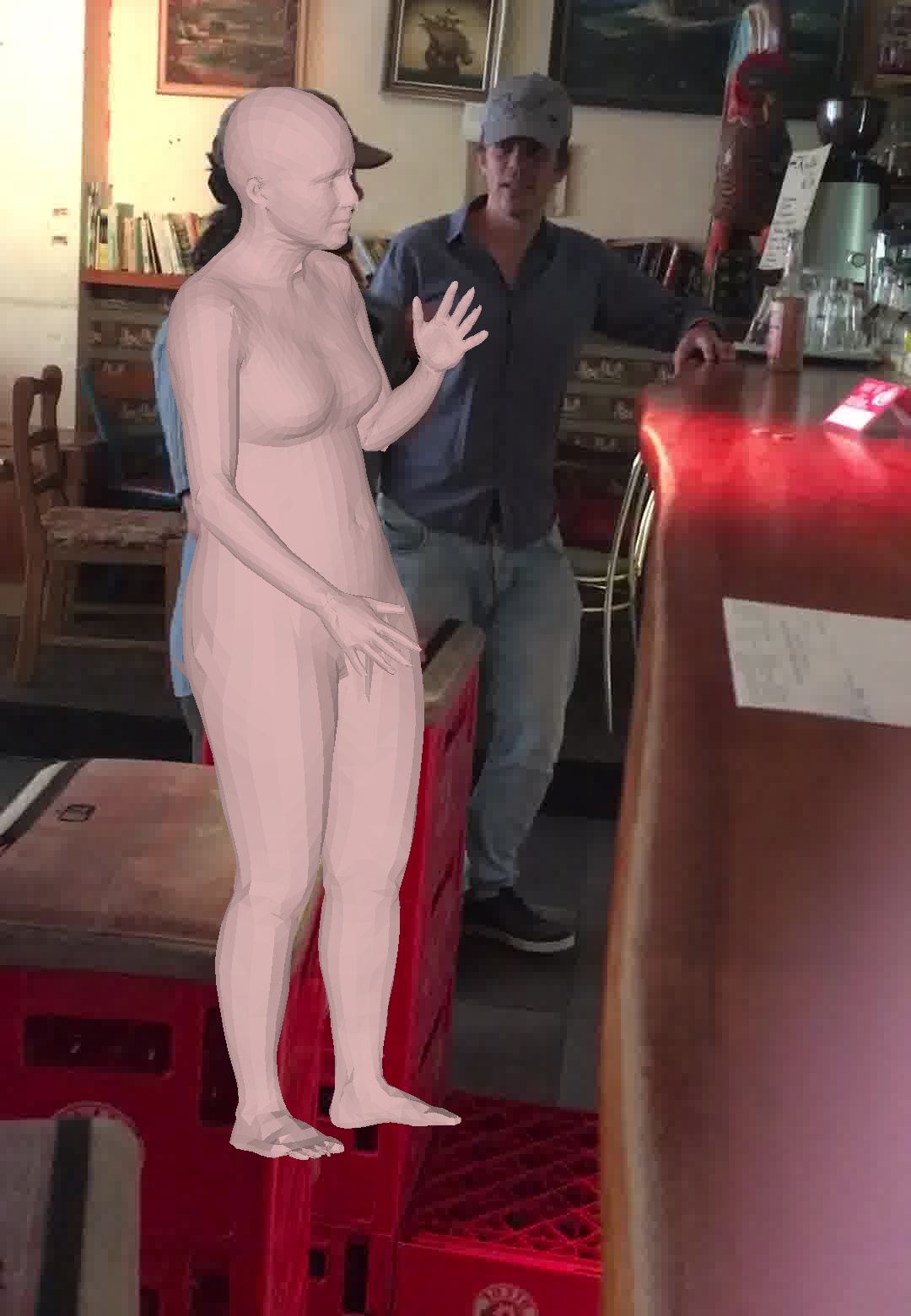} & 
\includegraphics[height=20mm]{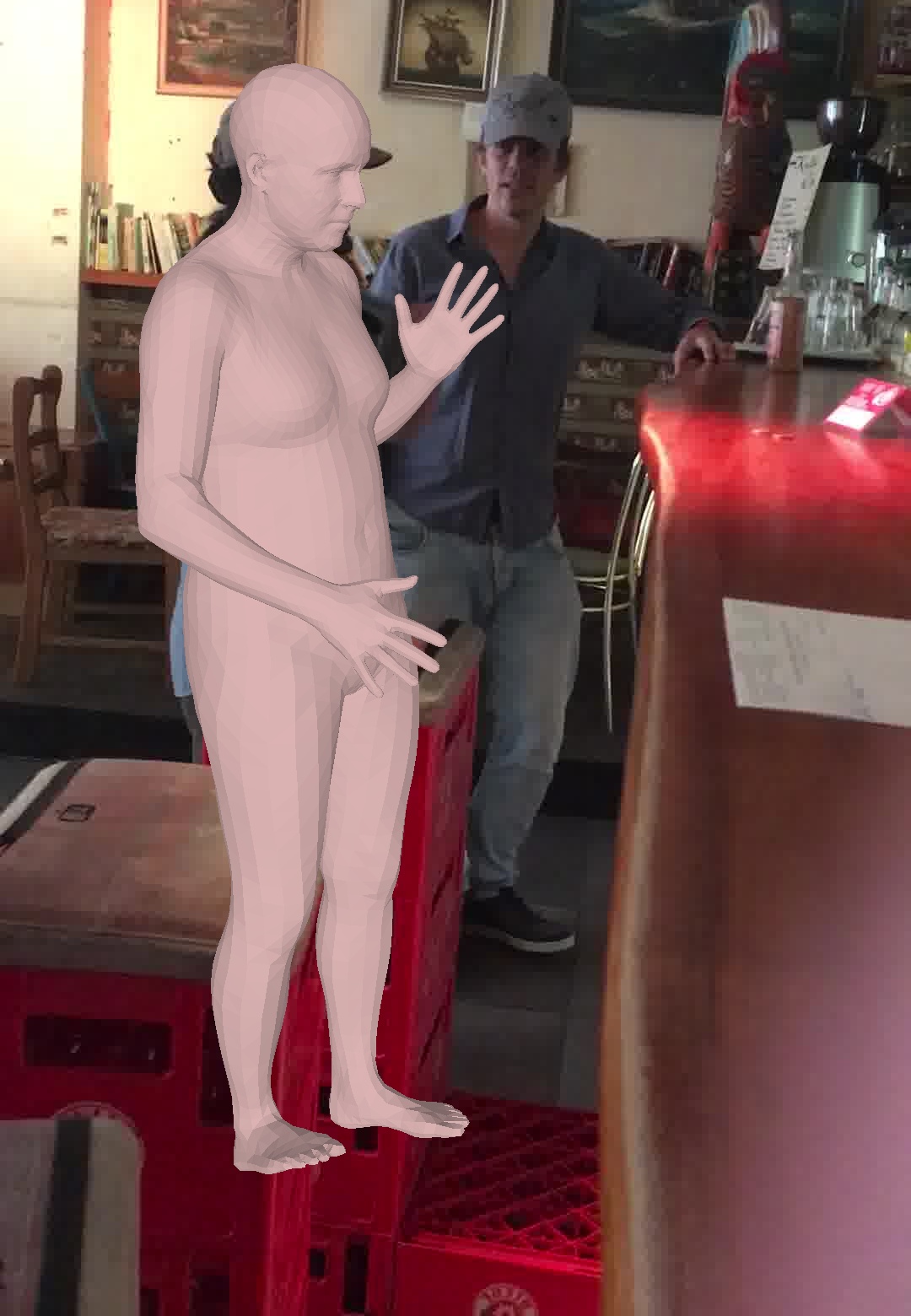} & 
\includegraphics[height=20mm]{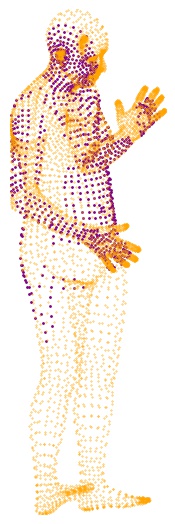} & 
\includegraphics[height=20mm]{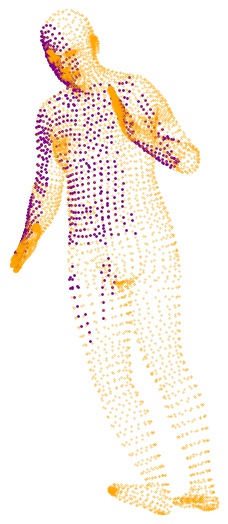}
\\
\includegraphics[height=20mm]{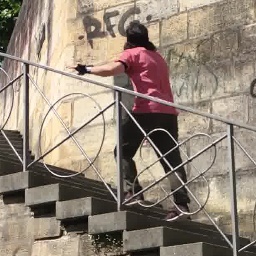} &
\includegraphics[height=20mm]{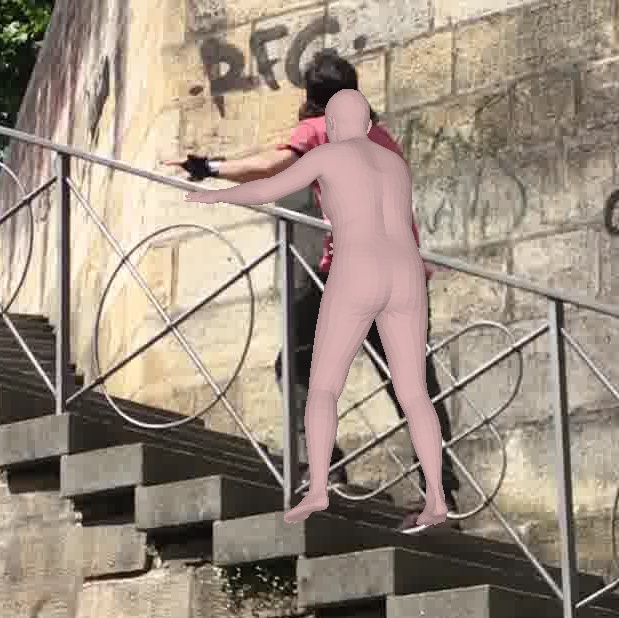} & 
\includegraphics[height=20mm]{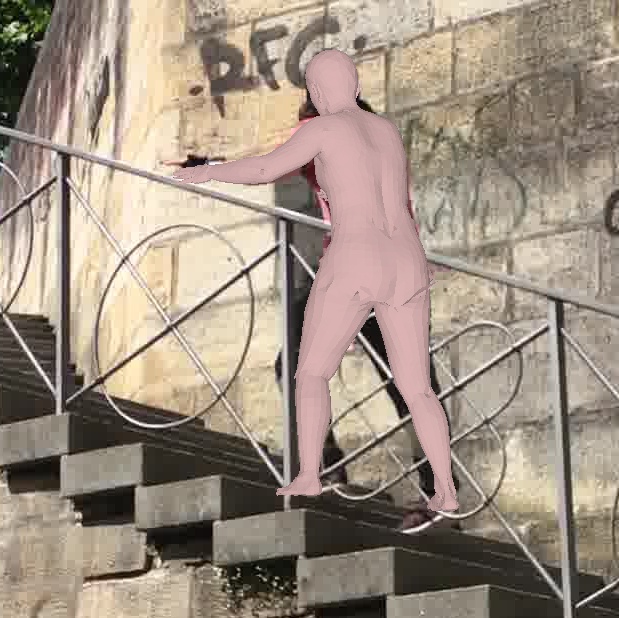} & 
\includegraphics[height=20mm]{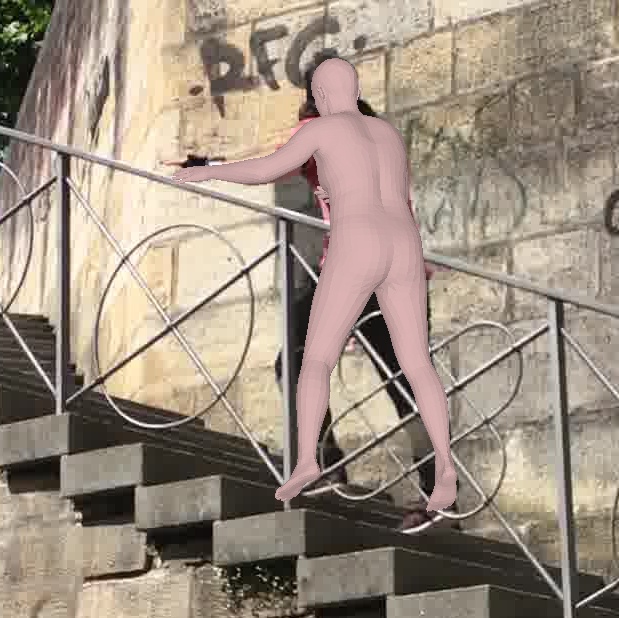} & 
\includegraphics[height=20mm]{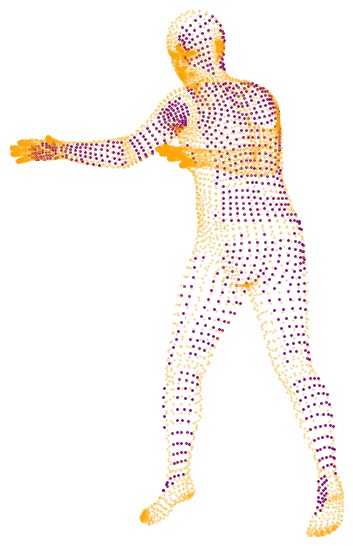} & 
\includegraphics[height=20mm]{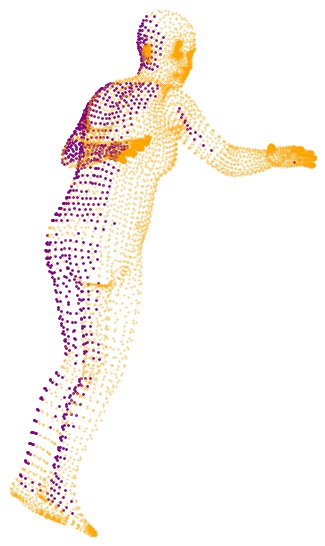}
\\
\includegraphics[height=20mm]{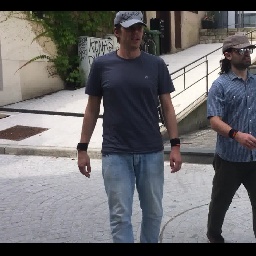} & 
\includegraphics[height=20mm]{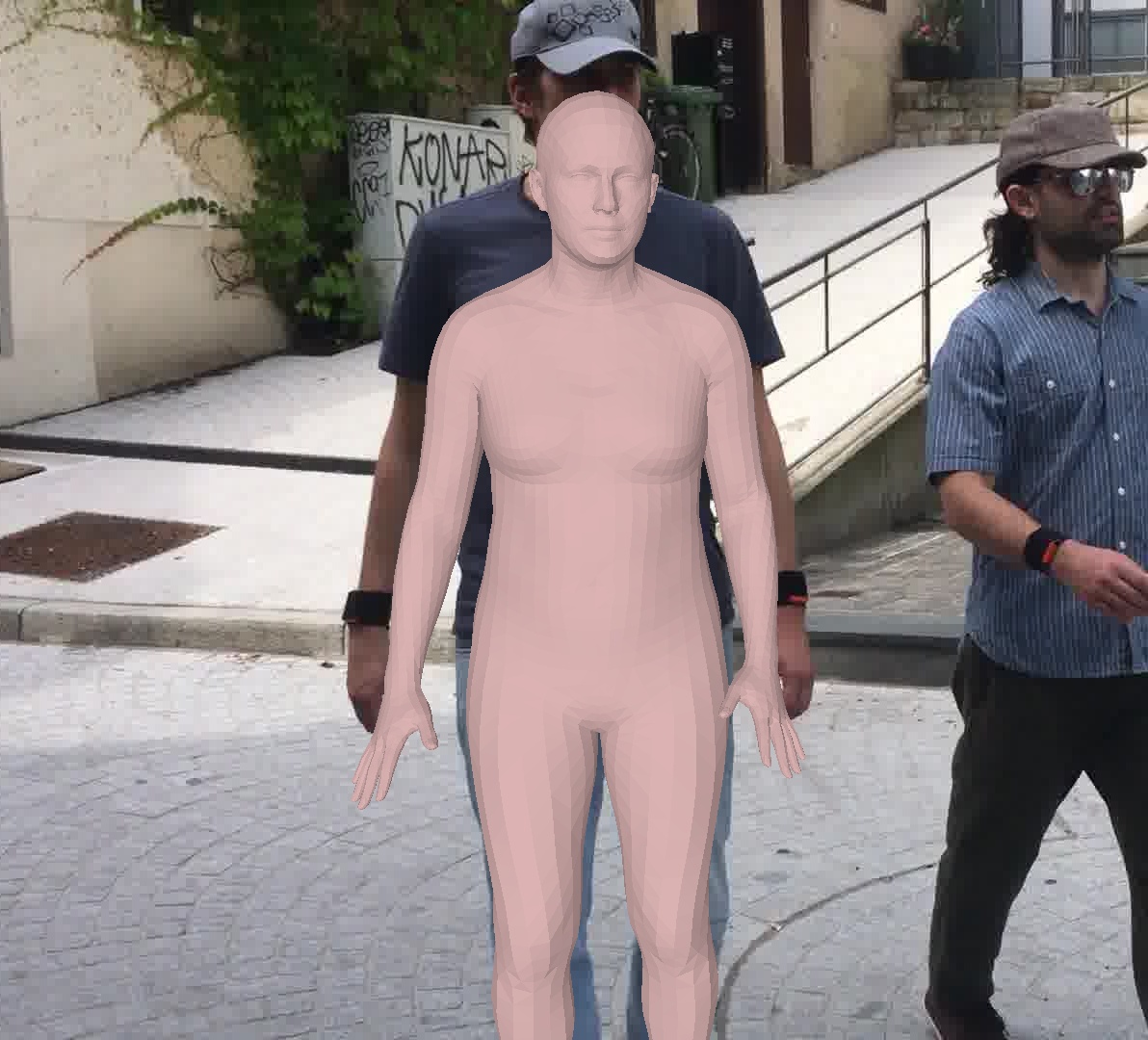} & 
\includegraphics[height=20mm]{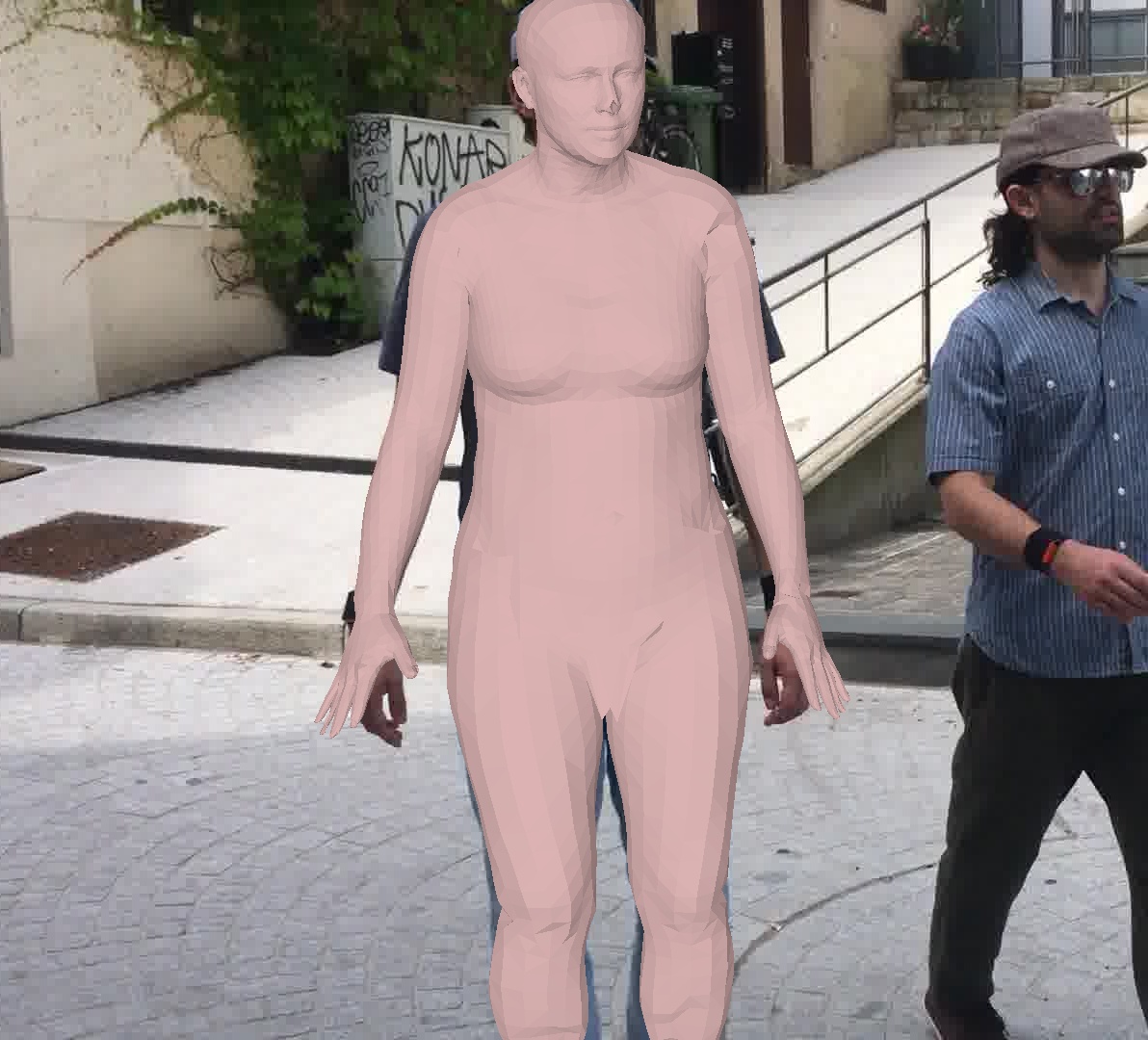} & 
\includegraphics[height=20mm]{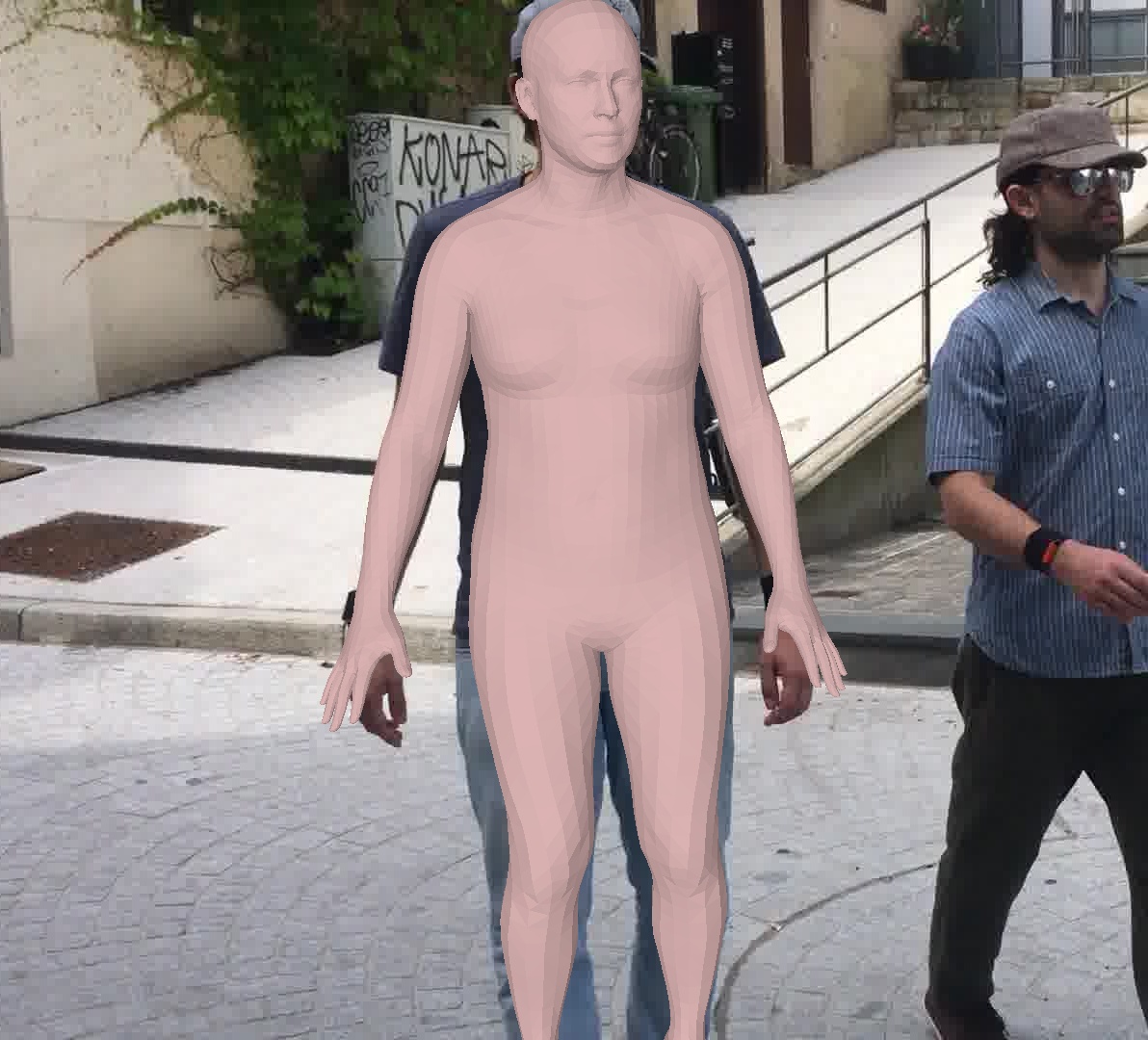} & 
\includegraphics[height=20mm]{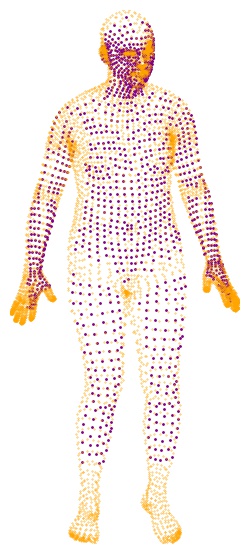} & 
\includegraphics[height=20mm]{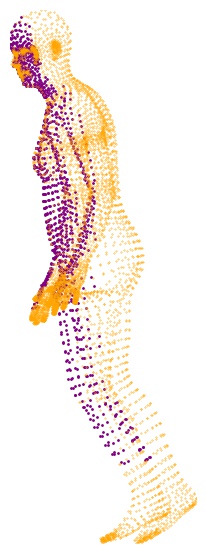}
\\
\includegraphics[height=20mm]{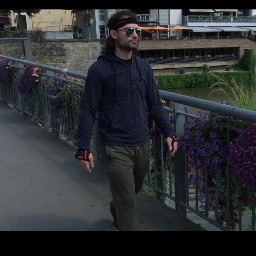} & 
\includegraphics[height=20mm]{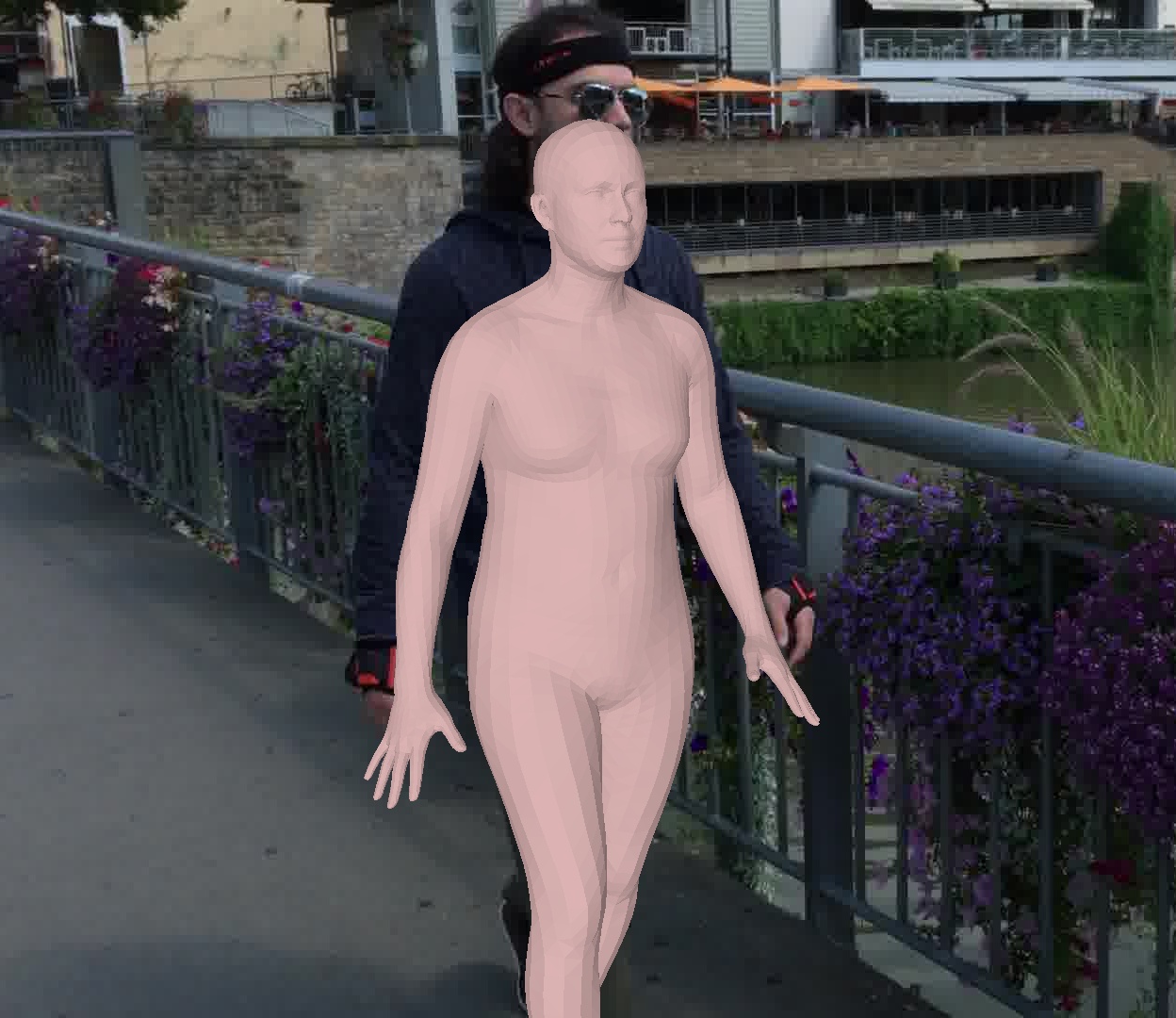} & 
\includegraphics[height=20mm]{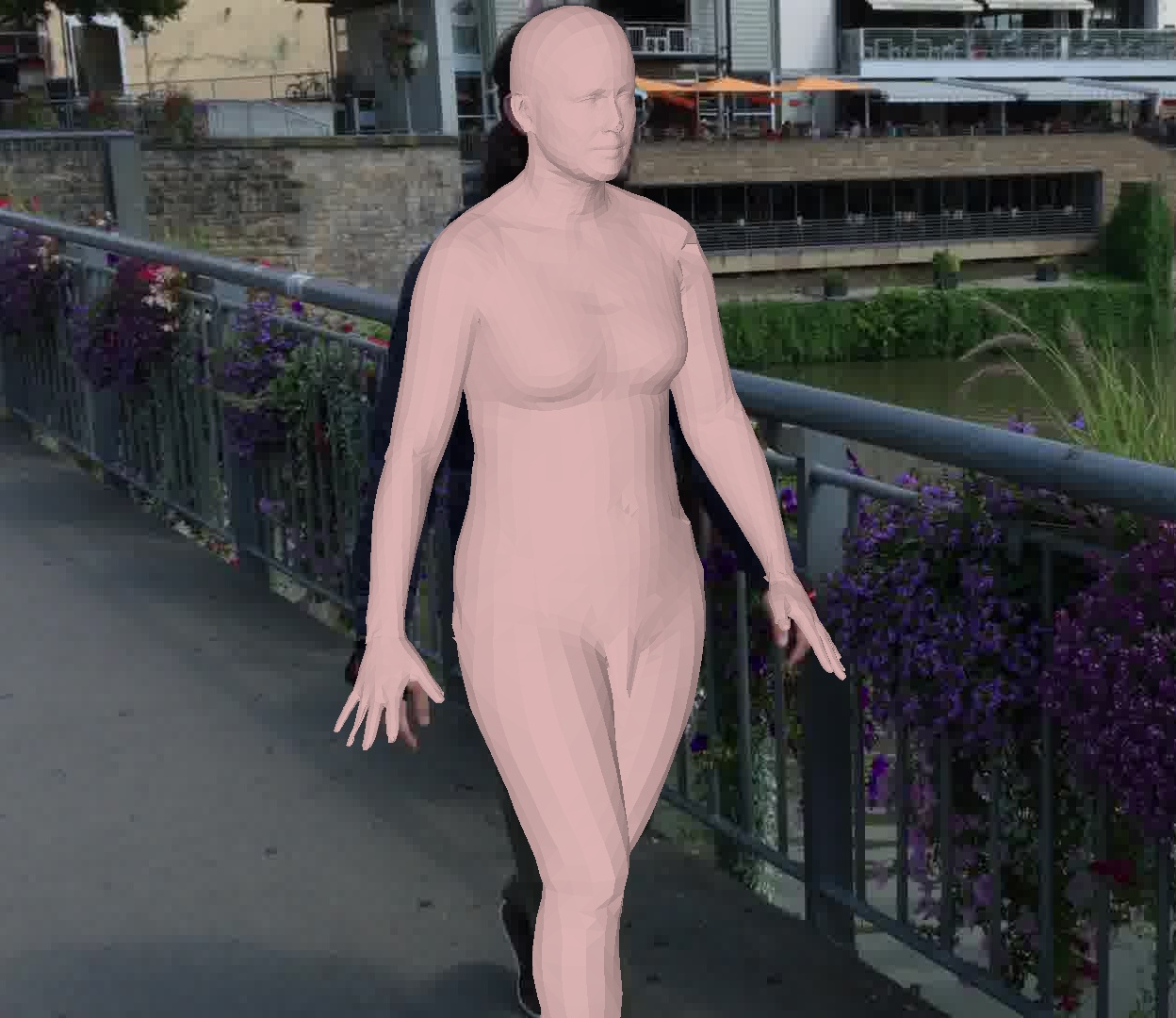} & 
\includegraphics[height=20mm]{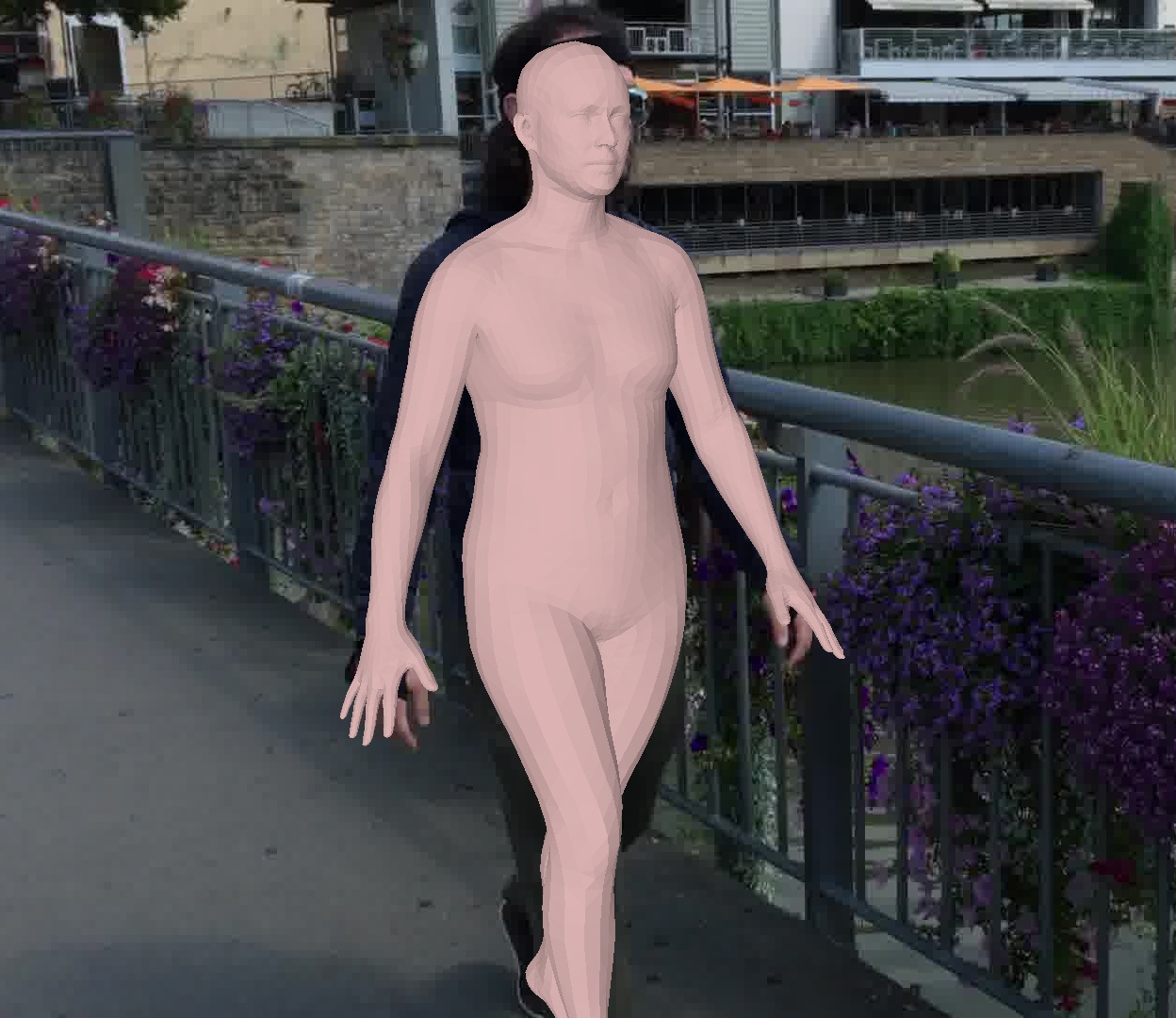} & 
\includegraphics[height=20mm]{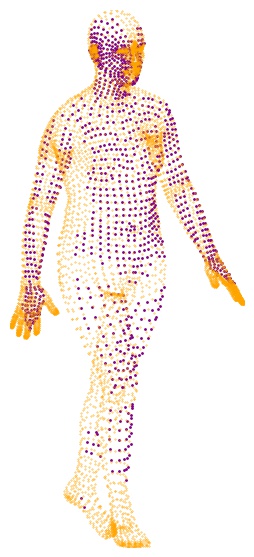} & 
\includegraphics[height=20mm]{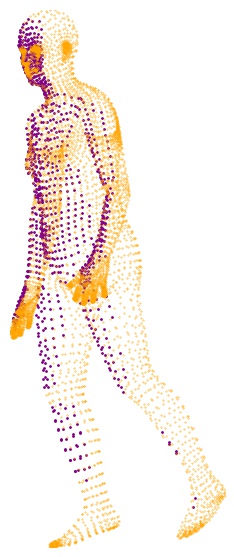}
\\
Input & I2L-MeshNet~\cite{moon2020i2l} & VisDB  & VisDB   & Visibility   & Visibility 
\\
      & (param)                        & (mesh) & (param) & (front view) & (side view)
\end{tabular}
%
\caption{\textbf{Qualitative results on the 3DPW dataset~\cite{von2018recovering}.} For each example, we show the results of I2L-MeshNet~\cite{moon2020i2l} SMPL model, our VisDB mesh, our optimized SMPL model, as well as visibility predictions in the front and side views (\textcolor{purple}{purple:visible}, \textcolor{orange}{orange:invisible}). When the human body is occluded (top two rows) or truncated (bottom two rows), both our VisDB output and optimized SMPL mesh capture the human silhouettes faithfully (\eg, the left hand in row 1 and the head region in rows 2,3,4).}
\label{fig:results}
\end{figure*}

\subsection{Qualitative Results}
%
Figure~\ref{fig:results} shows sample results by VisDB and I2L-MeshNet~\cite{moon2020i2l} on the 3DPW dataset~\cite{von2018recovering}.
I2L-MeshNet~\cite{moon2020i2l} regresses SMPL parameters from the entire heatmap-based mesh output, which leads to erroneous output meshes on truncated or occluded examples.
VisDB predicts accurate vertex visibility labels, improving both the image-space dense body estimation and SMPL parameter optimization.
The results show that VisDB (mesh) outputs can fit the human silhouettes faithfully, and VisDB (params) further regularizes and smooths the mesh surfaces.
More qualitative results are shown in the supplemental material.

\section{Conclusions}

In this work, we address the problem of dense human body estimation from monocular images.
Particularly, we identify the limitations of existing model-based and heatmap-based representations on truncated or occluded bodies.
As such, we propose a visibility-aware dense body representation, VisDB.
We obtain visibility pseudo ground-truths from dense UV correspondences and train a network to predict 3D coordinates as well as truncation and occlusion labels for each human joint and vertex.
Extensive experimental results show that visibility modeling can facilitate human body estimation and allow accurate SMPL fitting from partial-body predictions.

\clearpage
%
%
\bibliographystyle{splncs04}
\bibliography{egbib}
\end{document}